\documentclass{article}
  
\usepackage[final]{arxiv}
\usepackage{hyperref}
\usepackage{graphicx}
\usepackage{algorithm, algpseudocode}
\usepackage{amsmath, amsfonts, amsthm, amssymb}

\title{Reinforcement Learning with A* and a Deep Heuristic}

\author{
  Ariel Kesleman\\
  Imagry\\
  \href{mailto:ariel@imagry.co}{ariel@imagry.co}\\
  \And
  Sergey Ten\\
  Imagry\\
  \href{mailto:}{sergey@imagry.co}\\
  \And
  Adham Ghazali\\
  Imagry\\
  \href{mailto:adham@imagry.co}{adham@imagry.co}\\
  \And
  Majed Jubeh\\
  Imagry\\
  \href{mailto:majed@imagry.co}{majed@imagry.co}\\
}

\newcommand{\alephstar}{$\aleph^*$ }
\newcommand{\astar}{$A^*$ }

\begin{document}

\maketitle

\begin{abstract}
\astar is a popular path-finding algorithm, but it can only be  applied to those domains where a good heuristic function is known. Inspired by recent methods combining Deep Neural Networks (DNNs) and trees, this study demonstrates how to train a heuristic represented by a DNN and combine it with \astar. This new algorithm which we call \alephstar can be used efficiently in domains where the input to the heuristic could be processed by a neural network. We compare \alephstar to N-Step Deep Q-Learning (DQN \citealt{mnih2013playing}) in a driving simulation with pixel-based input, and demonstrate significantly better performance in this scenario.
\end{abstract}

\section{Introduction}
\label{sec:introduction}

The problem of finding the minimal cost path between two nodes on a graph can be formulated as a decision process where at every visited node an optimal action has to be taken minimizing the total accumulated cost. Replacing cost by reward, any such algorithm that generates cost minimizing actions generates reward maximizing actions thus becoming a candidate solver for Markov Decision Processes (MDPs). \astar \citep{hart68} is such a cost minimization algorithm that takes advantage of domain knowledge in the form of a heuristic function; it is interesting because for certain conditions (admissibility and consistency of the heuristic) \astar converges to the optimal solution while visiting a minimal number of nodes, i.e. no other similar algorithm (equally informed) could perform better. 

The strength of \astar is also its main weakness: when a heuristic is unknown other algorithms are used, most notably Monte Carlo Tree Search (MCTS, \citealt{abramson1987}). MCTS randomly samples the configuration space and balances exploration vs exploitation. Most modern implementations are based on the Upper Confidence bound applied to Trees (UCT, \citealt{kocsis2006}). MCTS is a rollout based method, nodes are searched by walking the tree from the root, leafs are evaluated by performing (possibly) random actions until the episode ends. MCTS was successfully used for games like go since the early 90's \citep{brugmann1993, gelly2006}, and more recently with greater success by replacing the random rollout with a deep convolutional network \citep{silver2016, anthony2017, silver2017, silver2017mastering}.

Following the recent work with MCTS, in this study we address the main weakness of \astar by replacing the rule-based heuristic with a Neural Network, specifically by a Convolutional Neural Network (CNN), resulting in the algorithm \alephstar. The weights are learned in a Reinforcement Learning (RL) fashion, interacting with a simulated environment by performing actions and earning rewards. The algorithm is model-based, episodes are not linear, all visited states and their possible actions are kept in a priority queue for later consideration. Transition of states are kept in a tree structure\footnote{ideally should be a directed graph since different actions could lead to the same state, but it is not obvious how much this would help in practice}. Action values $Q$ are backpropagated along the tree to satisfy a variant of time-difference equation. Unlike MCTS the backpropagation happens only once the tree is completed and for all leafs at once. And unlike a common time-difference equation which use maximum of Q over actions we choose to use a variant of ``soft" maximum:
\begin{equation}
  \label{eq:backprop}  
  Q^s_a = r^s_a + \gamma \frac{
    \sum_{b\in{\mathcal A}} {Q^x_b w^x_b}
  }{
    \sum_{b\in{\mathcal A}} {w^x_b}
  }\;.
\end{equation}
Here $\gamma$ is the discount factor, $r^s_a$ denotes the reward given by performing action $a$ on state $s$, ${\mathcal A}$ denotes the set of possible actions which without loss of generality is assumed to be similar across all possible states, $x$ denotes the state resulting from taking action $a$ on state $s$, and $w$ is an action value weighting factor as explained in sec. \ref{sec:algorithm}. The heuristic ${\mathcal H}_\theta$ is represented by a deep neural network with parameters $\theta$ and taking as input the sensors $S$, themselves a function of state\footnote{this formalism allows accommodating different sensors per state}. The heuristic is trained to predict the action values by using some loss function, in the reference implementation we use 
\begin{equation}
  \mathcal{L} = \frac{\left| Q^s - {\mathcal H}_\theta(S(s)) \right|^2}{{\rm len}({\mathcal A})}
\end{equation}
where $Q^s$ is a vector of action values for state $s$ with an entry for every possible action $a \in \mathcal{A}$.

In our experiments the resulting tree is very efficient, it has little branching, a property enabling its use in runtime. The reason for this is that while \alephstar uses random exploration during training, it uses pure exploitation during evaluation. Whether this is better or worse than UCT depends of course on the quality of the heuristic, and remains an open question.

The structure of this paper is as follows: in section \ref{sec:algorithm} we describe in detail the \alephstar algorithm and a reference implementation, in section \ref{sec:experiment} we describe the experiment and environment used in this study, in section \ref{sec:results} we present the main results together with an explanation on reproduction, and we conclude with a summary in section \ref{sec:summary}. 

\section{The Algorithm}
\label{sec:algorithm}

\alephstar is a general model-based reinforcement learning algorithm and it uses the regular notion of an agent interacting with a simulated environment. A reference implementation of the algorithm described here can be found at \url{https://github.com/imagry/aleph_star}.

The training algorithm consists of building randomly initialized trees as described in Alg. \ref{alg:gentree}. The basic building block of the tree is a node, which contains the necessary information to maintain the tree structure (pointers to its parent and children), and also important information to perform the backpropagation of the $Q$ values from leafs to root according to the weighted time-difference as required by equation \ref{eq:backprop}. The backpropagation of action-values happens only once the tree is completed and it is done for all leafs at once. Adding new nodes, choosing actions does not require a ``rollout" walking from the root of the tree, instead possible actions and their parent nodes are added to and retrieved from a priority queue maximizing $C+Q_a$ where $C$ is the (non-discounted) total accumulated reward. Actions chosen for expansion not only maximize future discounted reward like in DQN, but they take into consideration also actual reward. After a tree is backpropagated, the experiences are stored in an experience buffer, which is then used to update the heuristic weights $\theta$ using any gradient descend method. The experience buffer is implemented as an array of tuples $[(Q^s, S(s)), \ldots]$ each containing a vector of action values and the sensors for the same state. It is built by iterating over tree nodes, calculating the sensors, and pushing new tuples to the array.

The weights $w^x_b$ in eq. \ref{eq:backprop} are the number of sub-nodes explored following action $b$. This results in larger weights for more thoroughly explored actions. For a good heuristic most propagation would happen through a single main tree ``trunk" as it would dominate weights. In the extreme this just converges to N-Step DQN. For completely unexplored nodes we set $w=1$ for all actions. Other weighting schemes could be used, for eg.g setting $w=0$ and $w_{argmax(Q)}=1$ which propagates through action-value maximizing branches (classical Q-Learning), a soft-max strategy, or weights that depend on the depth of the nodes. Also a threshold to $w$ could be applied, but we find that our simple strategy for weighting works good enough. Backpropagating according to Eq. \ref{eq:backprop} is easy to implement if child nodes are linearly added to the tree array always after their parent: iterating in reverse guarantees that all children of the current node were previously visited.

Finally, after each training iteration the exploration parameter $\epsilon$ and the learning rate can be updated just like in DQN. Many of the improvements applied to DQN can be implemented for \alephstar too: the experience replay can include a priority based on loss \cite{schaul2015prioritized} (as implemented in accompanying code), exploration rewards based on curiosity can be added \cite{burda2018exploration}, rewards can be clipped \cite{pohlen2018observe}, a target network can be used, and so on. In the reference implementation instead of using a target network we remove nodes that are too old from the experience buffer. This has the advantage of evaluating the heuristic only once per experience. Another detail regards the implementation of the priority queue: for efficient insertion and removal of values it is implemented using a heap. Efficiently popping random values from a heap is hard to do, hence instead of deleting values we just tag them as used. When the ratio of number of used to unused values crosses a threshold a garbage-collection cycle is run, it rebuilds the data-structure without the tagged entries.

For the runtime algorithm the heuristic can be used by itself just by calculating the sensorsy input, evaluating the heuristic and performing the action that maximizes the action value. Another option is to create a small tree guided by the heuristic with $\epsilon=0$, and acting on the action that leads from the root node to the last added node. When $\epsilon=0$ the last added node maximizes $Q_a + \mathit{node.C}$ i.e. accumulated reword plus ``past" reward. This is different than DQN or MCTS which just maximize $Q$. If the heuristic is good, the tree can be much thinner than an MCTS, allowing even small trees to be highly effective.

Note that Alg. \ref{alg:gentree} assumes infinite long episodes i.e. ending an episode is always considered a bad thing getting the minimal value possible of $0$. This requires always positive rewards, and positive heuristic. This heuristic behavior can be enforce in numerous ways, for e.g. by passing the output through an $\rm abs()$ non-linearity at the end of the neural network. Alg. \ref{alg:gentree} also assumes a minimum of two nodes per tree, as it is arguable whether a smaller tree can still be called a tree.

\begin{figure}
  \centering
  \includegraphics[width=1.0\textwidth]{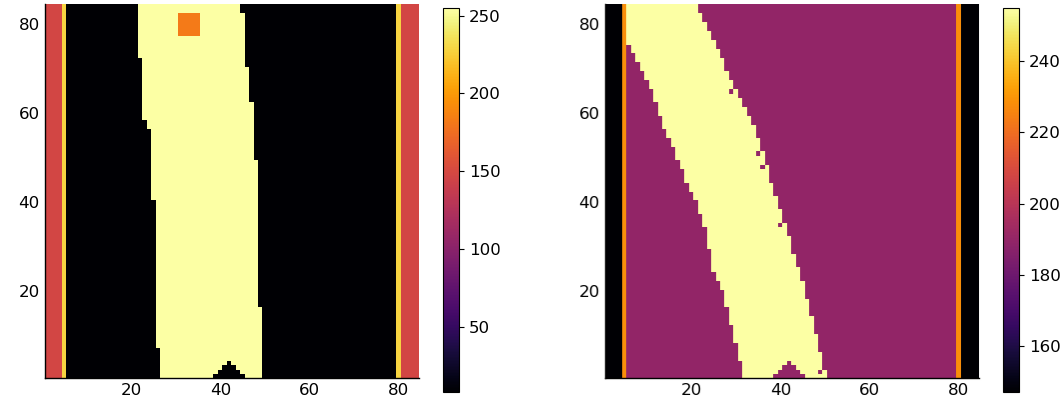}
  \caption{Two sampled sensors from different environment initialization. The sensors include a road, the actor car, other cars, and some information encoded in the pixel values as described in the text. The sensors are grayscale, a color palette is used here for better presentation. }
  \label{fig:sensors}
\end{figure}

\begin{algorithm}[h]
  \caption{Generate a tree guided by the heuristic ${\mathcal H}_\theta$}
  \label{alg:gentree}
  \begin{algorithmic}[1]
    \State initialize an empty vector of nodes $\mathit{Tree}$
    \State initialize a priority queue $\mathit{Queue}$
    \State {\bf initialize the root node:}
    \State $\mathit{node} \leftarrow$ new empty node
    \State $\mathit{node.s} \leftarrow$ randomly initialized state
    \Comment save the initial state in the root node
    \State $\mathit{node.C} \leftarrow 0$
    \Comment accumulated (past) reword is zero for root
    \State $\mathit{node.R} \leftarrow 0$
    \Comment reward given by performing action leading to this root node
    \State $\mathit{node.Done} \leftarrow$ False
    \Comment root is a non-terminal state
    \State $\mathit{node.Q} \leftarrow 0)$
  \Comment vector of action values, one entry per action  
  \State $\mathit{node.Parent} \leftarrow$ None
  \Comment root node has no parent
  \State $\mathit{node.Children} \leftarrow \{ \}$
  \Comment a dictionary actions $\rightarrow$ children nodes
  
  \State $\mathit{Tree} \leftarrow \mathit{node}$
  \Comment append root node to tree
  
  \State {\bf repeatedly add nodes until tree is large enough:}
  \Repeat
  \State $\mathit{node.Q} \leftarrow {\mathcal H}_\theta(S(s))$
  \If{not $\mathit{node.Done}$}
  \ForAll {$a \in \mathcal A $}    
  \State $\mathit{Queue} \leftarrow (a, \mathit{node})$ with priority $\mathit{node.C+\mathit{node.Q_a}}$
  \EndFor
    \EndIf
    \If{$rnd() < \epsilon$}
      \Comment exploration parameter
      \State $a, node \leftarrow \mathit{Queue.popRand()}$
    \Else
      \State $a, node \leftarrow \mathit{Queue.popMax()}$
    \EndIf
    \State $\mathit{newState, reward, done} \leftarrow \mathit{simulate(node.s, a)}$
    \State $\mathit{newNode} \leftarrow$ new empty node
    \State $\mathit{newNode.s} \leftarrow \mathit{newState}$
    \State $\mathit{newNode.C} \leftarrow \mathit{node.C + reward}$
    \Comment no discount for past accumulated reward
    \State $\mathit{newNode.R} \leftarrow \mathit{reward}$
    \State $\mathit{newNode.Done} \leftarrow \mathit{done}$
    \State $\mathit{newNode.Q} \leftarrow 0$
    \State $\mathit{newNode.Parent} \leftarrow \mathit{node}$
    \State $\mathit{node.children[a]} \leftarrow \mathit{newNode}$

    \State $\mathit{Tree} \leftarrow \mathit{newNode}$
    \Comment append node to tree    

    \State $\mathit{node} \leftarrow \mathit{newNode}$
  \Until{len($\mathit{Queue}$) $==$ 0 or len($\mathit{Tree}$) $>=$ maximum number of nodes in tree}
  \State {\bf return} $\mathit{Tree}$
\end{algorithmic}
\end{algorithm}

\section{Experiment}
\label{sec:experiment}

We implemented a pixel-based driving environment, the sensors are represented as an 84x84 grayscale image, the heuristic architecture is mostly similar to the one used by \cite{mnih2013playing}, it consists of a convolution layer of 16 8x8 filters with stride 4 followed by a non-learnable layer normalization \citep{ba2016layer} and a leaky-relu non-linearity \citep{xu2015empirical} with $\alpha=0.3$. The second convolutional layer consists of 32 4x4 filters with stride 2 similarly followed by layer normalization and leaky-relu. The last hidden layer is fully connected with 256 linear units, a layer normalization, and leaky-relus. The output layer consists of 35 linear units, as the number of actions in $\mathcal{A}$ (7 steering angles and 5 accelerations).

The sensors can be seen in Fig. \ref{fig:sensors}. Instead of using multiple frames as input, and since the simulation was developed by us, we chose to encode temporal information (the information affecting system dynamics such as velocity and steering angle) in the values of pixels: car velocity relative to the road is encoded in the background and in the color of the car itself. Relative velocity to other cars is encoded in the color of the target cars; and finally steering angle is encoded in the color of the vertical edges of the image.

The reward used is proportional to the car velocity, and multiplied by a factor preferring central positions and angles tangential to the lane. A small constant ``keep-alive" reward is also added every simulated time-step. The heuristic was trained with SGD, learning rate kept constant 0.01, and gamma 0.98. Batch size 64, and $\epsilon$ changed from an initial value of 0.5 to 0.01. Maximum tree size was 5500 nodes, and training continued for 1000 iterations. The N-Step DQN implementation closely follows \alephstar, except there is no priority queue, and branching is not allowed. There are plenty more details, all are present in the reference implementation.

\section{Results}
\label{sec:results}

\begin{figure}
  \centering
  \includegraphics[width=1.0\textwidth]{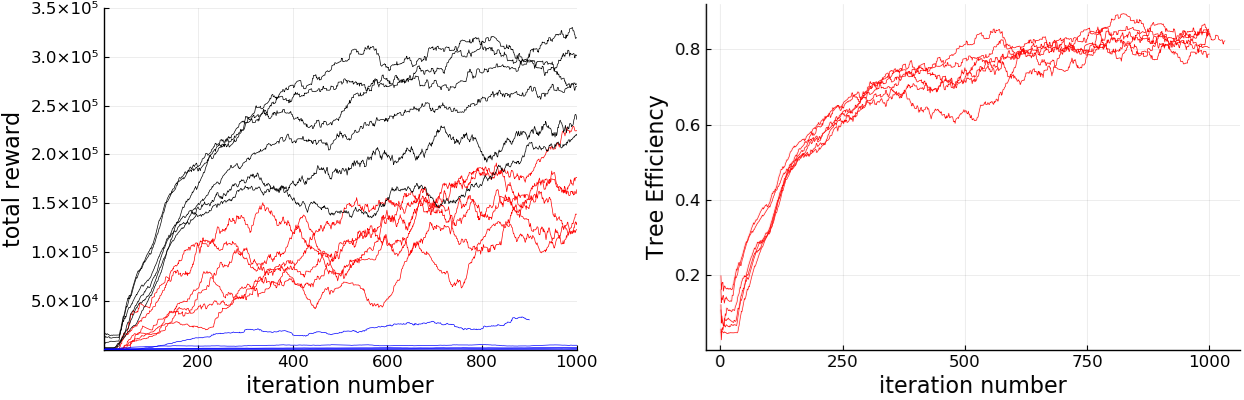}
  \caption{Main results for a few randomly initialized environments. Left panel accumulated reward as function of training iterations with a moving average window of 50 (black for the 5500 node tree, red for the heuristic alone, and blue for N-Step DQN), right panel tree efficiency as defined in the text}
  \label{fig:results}
\end{figure}

The results are summarized in Fig. \ref{fig:results}. Looking at the left panel. Training of the heuristic using \alephstar was efficient at 1000 iterations, reaching up to 50\% of the theoretical upper bound cumulative reward (given maximum number of nodes). Using the heuristic alone without a tree resulted in 50\% performance reduction (down to 25\% of the theoretical maximum) and N-Step DQN was not able to learn effectively. We also tried a few runs of N-Step DQN starting with $\epsilon=1$ but results were similar.

We define the rank of an \alephstar tree as the depth of the node maximizing accumulated reward $C$ + max($Q$). The efficiency is defined as the ratio of rank to maximal tree size. The efficiency is plotted in the right panel of Fig. \ref{fig:results}: the \alephstar tree, when trained, has little branching, and almost 85\% of the nodes are sequential. This means that a tree of only 10 nodes could, on average, be used for planning to a depth of 8 time-steps. It could be feasible to run such a tree on run-time. In the experiment trees regularly grow beyond a rank of 4000 while training. Since node choice is being done by a heap powered priority queue, there is no need to walk all these nodes in order to choose the next node to explore. Instead pushing or popping a single node from the queue takes constant time $\mathcal{O}(1)$ (for e.g. using a Fibonacci heap see \citealt{fredman1987fibonacci}), and hence building such tree scales as $\mathcal{O}(rank)$. In contrast, rollout based algorithms like MCTS scale as $\mathcal{O}(rank^2)$, potentially becoming a bottleneck when reaching rank in the thousands.

\section{Summary}
\label{sec:summary}

We presented a new model based reinforcement learning algorithm that efficiently combines a tree and a learnable heuristic. This algorithm has roots in \astar which was proven to be optimal under certain conditions, and as discussed in section \ref{sec:results} it is not rollout based, and it is suitable for use with very deep trees. We open-sourced the code for reproducing all the results presented in this paper at \url{https://github.com/imagry/aleph_star}, which include tests of \alephstar and N-Step DQN (as a baseline) in a pixel based sensory context. In that environment \alephstar learns effectively while N-Step DQN fails to learn at all. A proper comparison to MCTS (or AlphaZero) is still to be done. Further study is also needed to understand performance in more environments for e.g. Atari (The Arcade Learning Environment, \citealt{bellemare2013arcade})


\newpage

{
\bibliographystyle{apalike}
\small
\bibliography{aleph_star}
}

\end{document}